\documentclass[letterpaper, 10pt, conference]{ieeeconf}

\IEEEoverridecommandlockouts                
\usepackage{subfigure}
\usepackage{algorithmic, algorithm}
\usepackage{graphicx}
\usepackage{wrapfig}
\usepackage{url}

\newcommand{\captionfonts}{\small}

\makeatletter  
\long\def\@makecaption#1#2{%
  \vskip\abovecaptionskip
  \sbox\@tempboxa{{\captionfonts #1: #2}}%
  \ifdim \wd\@tempboxa >\hsize
    {\captionfonts #1: #2\par}
  \else
    \hbox to\hsize{\hfil\box\@tempboxa\hfil}%
  \fi
  \vskip\belowcaptionskip}
\makeatother   

\title{A Real-Time Model-Based Reinforcement Learning \\Architecture for Robot Control}
\author{Todd Hester, Michael Quinlan, Peter Stone\\
The University of Texas at Austin\\
1616 Guadalupe, Suite 2.408\\
Austin, TX 78701\\
\texttt{\{todd,mquinlan,pstone\}@cs.utexas.edu}}

\begin{document}
\maketitle

\begin{abstract}
Reinforcement Learning (RL) is a method for learning
decision-making tasks that could enable
robots to learn and adapt to their situation on-line. For an RL
algorithm to be practical for robotic control tasks, it must learn in
very few actions, while continually taking those actions in
real-time. Existing model-based RL methods learn in relatively few
actions, but typically take too much time between each action for
practical on-line learning. In this paper, we present a novel parallel
architecture for model-based RL that runs in real-time by 1) taking
advantage of sample-based approximate planning methods and 2)
parallelizing the acting, model learning, and planning processes such
that the acting process is sufficiently fast for typical robot control
cycles. We demonstrate that algorithms using this architecture perform
nearly as well as methods using the typical sequential architecture
when both are given unlimited time, and greatly out-perform these
methods on tasks that require real-time actions such as controlling an
autonomous vehicle.
\end{abstract}

\section{Introduction}

Robots have the potential to solve many problems in society by working in dangerous places or performing jobs that no one wants. 
One barrier to their widespread deployment is the need to hand-program behaviors for every situation they may encounter. 
For robots to meet their potential, 
we need methods for them to learn and adapt to novel situations.

Reinforcement learning (RL)~\cite{sutton98reinforcement} is a method
for learning sequential decision making processes that could solve the
problems of learning and adaptation on robots. An RL agent seeks to
maximize long-term rewards through experience in its environment. The
decision making tasks in these environments are usually formulated as
Markov Decision Processes (MDPs).

RL has been applied to a few carefully chosen robotic tasks
that are achievable with limited training and infrequent action
selections (e.g.~\cite{kohl:aaai04}), or allow for an off-line learning
phase (e.g.~\cite{DBLP:conf/nips/NgKJS03}).
However, none of these methods allow for continual learning on the robot running in its environment.
For RL to be practical on tasks requiring lifelong continual control of a robot, such as low-level control tasks,
it must meet at least the following two requirements:
\begin{enumerate}
\item It must learn in very few actions (be \emph{sample efficient}).
\item it must take actions continually in real-time (even while learning).
\end{enumerate}

Model-based methods such as \textsc{r-max}~\cite{brafman01rmax} are a
class of RL algorithms that meet the first requirement by learning a
model of the domain from their experiences, and then planning a policy
on that model. By updating their policy using their model rather than
by taking actions in the real world, they limit the number of real
world actions needed to learn.

However, most existing model-based methods fail to meet the second
requirement because they take significant periods of
(wall-clock) time to update their model and plan between each action. 
These action times
are acceptable when learning in simulation or planning off-line,
but for on-line robot control learning, actions must be given at a
fixed, fast frequency. 
Some model-based methods that do take actions at this fast frequency
have been applied to robots in the past
(e.g.~\cite{DBLP:conf/nips/NgKJS03,Deisenroth2009}), 
but they perform learning off-line during pauses where they stop controlling the robot entirely.
\textsc{dyna}~\cite{Sutton90integratedarchitectures}, which does run in real-time, uses a simplistic model
and is not very sample efficient.
Model-free
methods also learn in real-time, but often take thousands of potentially
expensive or dangerous real-world actions to learn: they meet our
second requirement, but not the first.

The main contribution of this paper is a novel RL architecture, called
Real-Time Model Based Architecture (\textsc{rtmba}), that is the first to
exhibit both sample efficient and real-time learning, meeting
both of our requirements. It does so by leveraging recent
sample-based approximate planning methods, and most
uniquely, by parallelizing model-based methods to run in
real-time. With \textsc{rtmba}, the crucial computations needed to make
model-based methods sample efficient are still performed, but threaded 
such that actions are not delayed.
We compare \textsc{rtmba} with other methods in simulation when they
are all given unlimited time for computation between actions. We then
demonstrate that it is the only algorithm among them that successfully learns to control an autonomous vehicle, both in simulation and on the robot.

\section{Background}

We adopt the standard Markov Decision Process (MDP) formalism of
RL~\cite{sutton98reinforcement}. An MDP consists of a set of states
$S$, a set of actions $A$, a reward function $R(s,a)$, and a
transition function $P(s'|s,a)$. 
In each state $s \in S$, the agent takes an action
$a \in A$. Upon taking this action, the agent receives a reward
$R(s,a)$ and reaches a new state $s'$, 
determined from the probability distribution $P(s'|s,a)$.

The value $Q^*(s,a)$ of a state-action $(s,a)$ is an estimate of the expected long-term rewards that can be obtained from $(s,a)$ and is determined by solving the Bellman equation: 
\begin{eqnarray}
Q^*(s,a) = R(s,a) + \gamma \sum_{s'} P(s'|s,a) \max_{a'} Q^*(s',a')
\end{eqnarray}
where $0 < \gamma < 1$ is the discount factor.
The agent's goal is to find the policy $\pi$ mapping states to
actions that maximizes the expected discounted total reward over the
agent's lifetime.
The optimal policy $\pi$ is then: 
\begin{eqnarray}
\pi(s) = \textrm{argmax}_{a}Q^*(s,a)
\end{eqnarray}

Model-based RL methods learn a model of the domain by approximating $R(s,a)$ and $P(s'|s,a)$. The agent then computes a policy by planning on this model with a method such as value iteration~\cite{sutton98reinforcement}.
RL algorithms can also work without a model, updating action-values only when taking them in the real task.
Generally model-based methods are more sample efficient than model-free methods; their sample efficiency is only constrained by how many actions it takes to learn a good model.

\begin{figure}[tbp]
\begin{center}
\includegraphics[width=0.7\linewidth]{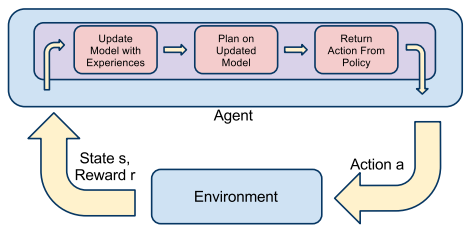}
\end{center}
\vspace{-0.2cm}
\caption{A diagram of how model learning and planning are typically interleaved in a model-based agent.}
\vspace{-0.3cm}
\label{fig:seq}
\end{figure}

Figure~\ref{fig:seq} shows the typical architecture for a model-based algorithm. When the agent gets its new state, $s'$, and reward, $r$, it updates its model with the new transition $\left<s,a,s',r\right>$. Once the model has been updated, it computes a new policy by re-planning on its model. The agent then returns the action for its current state determined by its policy. Each of these computations is performed sequentially and both the model learning and planning can take significant time. 

The \textsc{dyna} framework~\cite{Sutton90integratedarchitectures} presents an alternative to this approach. It incorporates some of the benefits of model-based methods while still running in real-time. \textsc{dyna} saves its experiences, and then performs $k$ Bellman updates on randomly selected experiences between each action. Instead of performing full value iteration between each action as above, its planning is broken up into a few updates between each action. However, it uses a very simplistic model (saved experiences) and thus does not have very good sample efficiency.
In the next section, we introduce a novel parallel architecture to
allow more sophisticated model-based algorithms to run in real-time 
regardless of how long the model learning or planning may take.

\section{The Architecture}
\label{sec:arch}

We make two main modifications to the standard model-based paradigm
that, together, allow it to run in real-time. First, we limit planning
time by using approximate instead of exact planning. Second, we
parallelize the model learning, planning, and acting such
that the computation-intensive processes (model learning and planning)
are spread out over time. Actions are produced as quickly as dictated
by the robot control loop, while still being based on the most recent
models and plans available.

First, instead of planning exactly with value iteration (like methods such as \textsc{r-max}), 
our method follows the approach of Silver et al.~\cite{ICML08-silver}
(among others) in using a sample-based planning algorithm from the Monte
Carlo Tree Search (MCTS) family (such as Sparse
Sampling~\cite{kearns99sparse} or
\textsc{uct}~\cite{Kocsis06banditbased}) to plan \emph{approximately}.
These sample-based planners perform rollouts from the agent's current
state, sampling ahead to update the values of the
sampled actions. The agent performs as many rollouts as it can in the
given time, with its value estimate improving with more
rollouts. These methods can be more efficient than dynamic programming
approaches in large domains because they focus their updates on states
the agent is likely to visit soon rather than iterating over the
entire statespace.

\begin{figure}[tbp]
\begin{center}
\includegraphics[width=0.7\linewidth]{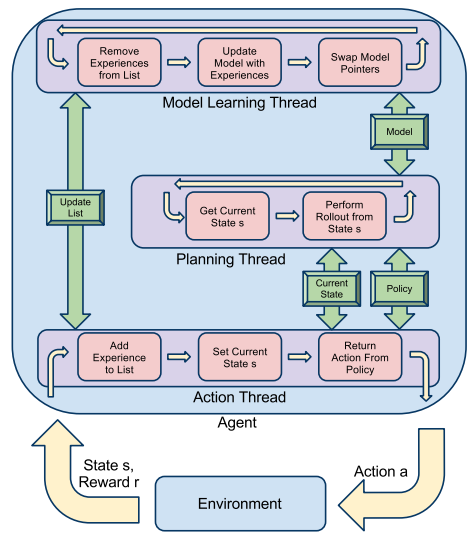}
\end{center}
\vspace{-0.2cm}
\caption{A diagram of the proposed parallel architecture for real-time model-based RL.}
\vspace{-0.3cm}
\label{fig:parallel}
\end{figure}

Second, since both the model learning and planning can take
significant computation (and thus also wall-clock time), we place both
of those processes in their own parallel threads in the background, 
shown in Figure~\ref{fig:parallel}. 
A third thread interacts with the environment, receiving the agent's new
state and reward and returning the action given by the agent's current
policy. 
By de-coupling this action thread from the time-consuming
model-learning and planning processes, \textsc{rtmba} releases the algorithm
from the need to complete the model update and planning between
actions. Now, it can return
an action immediately whenever one is requested by the environment.

For the three threads to operate properly, they must share
information while avoiding race conditions and
data inconsistencies. The model learning thread must know which new
transitions to add to its model, the planning thread must access 
the model being learned, the planner must know what state the agent is
currently at, and the action thread must access the policy
being planned. \textsc{rtmba} uses mutex locks to control access to these
variables, as summarized in Table~\ref{tab:mutex}.

\begin{table}[b]
  \vspace{-0.3cm}
  \small
  \tabcolsep 4pt
  \centering
  \begin{tabular}{|l|l|l|}
        \hline
Variable & Threads & Use \\
        \hline
Update List           & Action         & Store experiences to \\ 
                      & Model Learning & be updated into model \\
\hline
Current State         & Action         & Set current state \\
                      & Planning       & to plan from \\
\hline
Policy (by state)     & Action         & Update policy used \\
(Value Function)      & Planning       & to select actions \\
\hline
Model                 & Model Learning & Latest model \\
                      & Planning       & to plan on \\
        \hline
\end{tabular}
\vspace{-0.05cm}
  \caption{This table shows all the variables that are protected under mutex locks in the proposed architecture, along with their purpose and which threads use them.}
  \label{tab:mutex}
\end{table}

The action thread receives the agent's new state and reward, and adds the new transition experience, $\left<s,a,s',r\right>$, to a list to be updated into the model. It then sets the agent's current state for use by the planner and returns the action determined by the agent's policy. The update list and current state are both protected by mutex locks, and the agent's policy is protected by individual mutex locks for each state.

The model learning thread checks if there are any experiences in the update list to be added to its model. If there are, it makes a copy of its model, updates it with the new experiences, and replaces the original model with the copy. 
The other threads can continue accessing the original model while the copy is being updated.
Only the swapping of the models requires locking the model mutex.
After updating the model, the model learning thread repeats, checking for new experiences to add to the model. 

The model learning thread can incorporate any type of model learning
algorithm, such as a tabular model~\cite{brafman01rmax}, 
random forests~\cite{ICDL10-hester} (as used in this paper), 
or Gaussian Process regression~\cite{Deisenroth2009}. 
Depending on how long the model
update takes and how fast the agent is acting, the agent can add tens or hundreds of new experiences to its model at a time, or
it can wait for long periods for a new experience.
When adding many experiences at a
time, full model updates are not performed between
each individual action. In this case, the algorithm's sample
efficiency is likely to suffer compared to that of sequential methods,
but in exchange, it continues to act in real time.

The planning thread uses any MCTS planning algorithm to plan
approximately (we use a variant of \textsc{uct}). It 
retrieves the agent's current state and its sample-based planner
performs a rollout from that state. The thread repeats, continually 
performing rollouts from the agent's current state. 
With more rollouts, the algorithm's estimates of 
action values improve, resulting in more
accurate policies. Even if very few rollouts are performed from the
current state before the algorithm returns an action, many of the
rollouts performed from the previous state should have gone through
the current state (if the model is accurate), giving the
algorithm a good estimate of its true value.

The action thread returns actions in real-time. When an action is requested, 
the action thread only
has to add an experience to the update list, set the agent's current
state, and access the agent's policy to return an action. All of these
items are under mutex locks, but the update list is only used by the
model learning thread between model updates, the agent's current state
is only accessed by the planning thread between each rollout, and the
policy is under individual locks for each state. Thus, any given
state is freely accessible most of the time. When the planner does
happen to be using the same state the action thread wants, it releases
it immediately after updating the values for that state. 
In addition to enabling real-time action, this architecture enables
 the agent
to take full advantage of multi-core processors by running each thread on
a separate core.\footnote{Source code for the architecture is available at: \url{http://www.ros.org/wiki/rl-texplore-ros-pkg}.}

\section{Experiments}

To demonstrate the effectiveness of this architecture, we performed
experiments on two problems. Our first experiments measure the cost of
parallelization in terms of environmental reward compared to a
traditional sequential architecture. 
We use
the standard toy domain of mountain car, in which the simulated
environment can wait as long as necessary for the agent to return an
action (or it can execute actions as fast as the algorithm returns them). 
Our second set of experiments measure the performance gains
due to parallelization on an autonomous vehicle, where real-time
actions are absolutely necessary. We perform experiments, both in simulation
and on the robot, that show that existing sequential
approaches are not a viable option on this type of problem.

\subsection{Mountain Car}
\label{sec:mcar}

\begin{wrapfigure}{l}{0.41\linewidth}
\vspace{-0.5cm}
\begin{center}
\includegraphics[width=0.44\columnwidth]{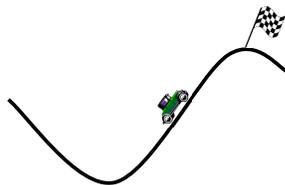}
\end{center}
\vspace{-0.2cm}
\caption{Mountain Car}
\vspace{-0.25cm}
\label{fig:mcar}
\end{wrapfigure}

Our first experiments were performed in the Mountain Car domain~\cite{sutton98reinforcement}, shown in Figure~\ref{fig:mcar}. Mountain Car is a continuous task, where the agent controls an under-powered car that does not have enough power to drive directly up the hill to the goal. Instead, it must go up the leftward slope to gain momentum first. The agent has three actions, accelerating it leftward, rightward, or not at all. The agent's state is made up of two features: its \textsc{position} and its \textsc{velocity}. The agent receives a reward of $-1$ each time step until it reaches the goal, when the episode terminates with a reward of $0$.
We discretized both state features into 100 values each, and ran the algorithms on the discretized version of the domain. 
Following the evaluation
methodology of Hester and Stone~\cite{ICDL10-hester}, each algorithm was initialized with one experience ($\left<s,a,s',r\right>$ tuple) of the car reaching the goal to jump start learning.

We ran experiments with a typical model-free RL method (\textsc{q-learning}~\cite{watkins89learning}), \textsc{dyna}, two sequential model-based methods, and \textsc{rtmba}. \textsc{dyna} performed updates on 1,000 experiences between each action. The sequential methods varied in their planning; one used value iteration for exact planning and one used \textsc{mcts} for approximate planning. We modified \textsc{mcts} to use \textsc{uct} action selection~\cite{Kocsis06banditbased}, eligibility traces, and to generalize values across depths in the search tree. 
Between each action, the two sequential methods performed a full model update, then planned on their model by running value iteration to convergence or running \textsc{mcts} for 0.1 seconds. 
We compared these algorithms with \textsc{rtmba} using the same version of \textsc{mcts}, running at three different action rates: 10 Hz, 25 Hz, and 100 Hz. All of the algorithms used random forests to model the domain, similar to the approach taken by Hester and Stone~\cite{ICDL10-hester}. We ran 30 trials of each algorithm learning for 1,000 episodes in the domain.
Each trial was run on a single core of a machine with 2.4 - 2.66 GHz Intel Xeon processors and 4 GB of memory. 

Our aim was to compare the real-time algorithms with the sequential methods
when they were given the time needed to fully complete their
computation between each step. Thus we can examine the performance
lost by the real-time algorithms due to acting quickly.
In contrast, the model-free methods could act as fast as they wanted, resulting in learning that took little wall clock time but many more actions.
To perform these experiments, the environment waited for each
algorithm to return its action. This is only possible in simulation, whereas on a real robot, the action rate is defined by the robot rather than the algorithm.

\begin{figure}[tbp]
\begin{center}
\includegraphics[width=0.9\linewidth]{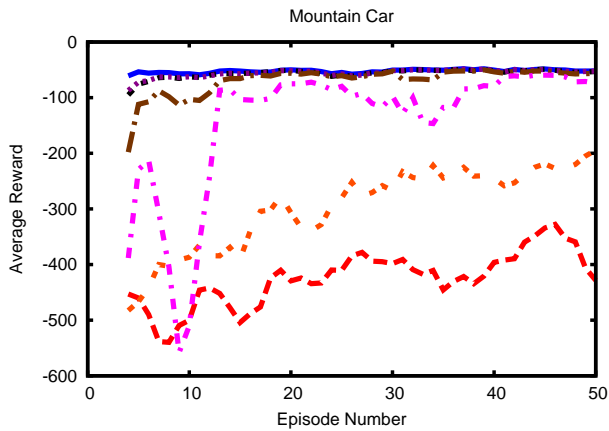}
\end{center}
\vspace{-0.2cm}
\caption{Average reward per episode on Mountain Car, averaged over 30 trials. Results are averaged over a 4 episode sliding window.}
\vspace{-0.3cm}
\label{fig:mcar-ep}
\end{figure}

\begin{figure}[tbp]
\begin{center}
\includegraphics[width=0.9\linewidth]{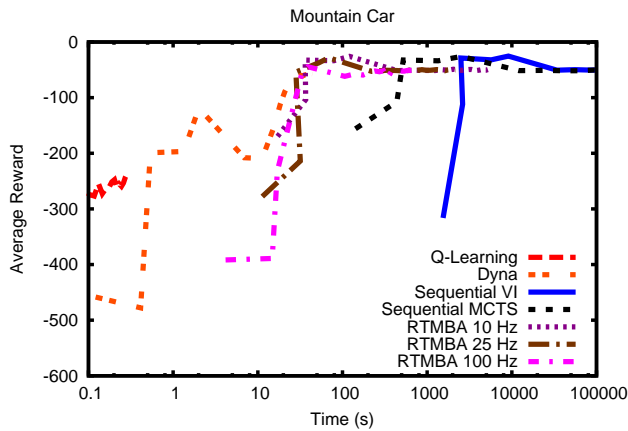}
\end{center}
\vspace{-0.2cm}
\caption{Average reward versus clock time on Mountain Car, averaged over 30 trials. Results are averaged over a 4 episode sliding window. Note that the $x$-axis is in log scale.}
\vspace{-0.3cm}
\label{fig:mcar-time}
\end{figure}

\begin{figure}[tbp]
\begin{center}
\includegraphics[width=0.9\linewidth]{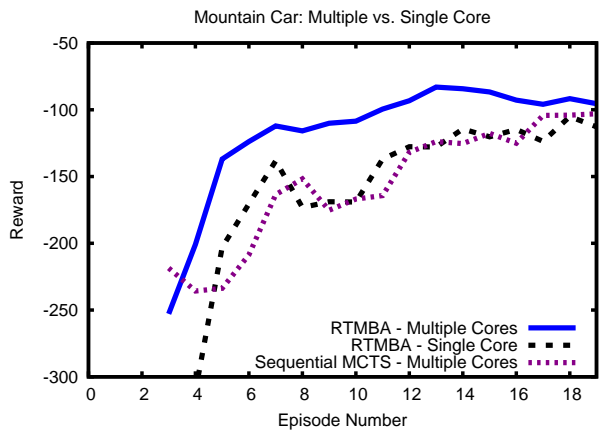}
\end{center}
\vspace{-0.2cm}
\caption{Comparisons of the methods using a multiple core machine. Each method is averaged over 30 trials on Mountain Car.}
\vspace{-0.3cm}
\label{fig:multivsingle}
\end{figure}

Figure~\ref{fig:mcar-ep} shows the average reward per episode for each algorithm over the first 50 episodes in the domain and Figure~\ref{fig:mcar-time} shows the reward plotted against clock time in seconds (note the log scale on the $x$ axis). The first plot shows that the two sequential methods perform better than \textsc{rtmba} in sample efficiency, in particular, receiving significantly more reward per episode than \textsc{rtmba} running at 25 and 100 Hz over the first 5 episodes ($p < 0.05$). \textsc{rtmba} running at 10 Hz did not perform significantly worse than the sequential method using \textsc{mcts}. However, Figure~\ref{fig:mcar-time} shows that better performance of the sequential methods came at the cost of more computation time. 
For the sequential methods, switching from exact to approximate planning reduces the time to complete the first episode from 1541 to 142 seconds, but the \textsc{mcts} method is still restricted by the need to perform complete model updates between actions. This restriction is removed with \textsc{rtmba}, and all three versions using it complete the first episode within 20 seconds. In fact, all three \textsc{rtmba} methods start performing well after 120 seconds, likely because they all took this much time to learn an accurate model of the domain. Compared with the sequential methods, \textsc{rtmba} is only slightly worse in sample efficiency, and is able to act much faster, meeting our second requirement of continual real-time action selection.

The two model-free approaches, \textsc{q-learning} and \textsc{dyna}, select actions extremely quickly and converge to the optimal policy in less wall clock time than any version of \textsc{rtmba}. However, Figure~\ref{fig:mcar-ep} shows that they are not as sample efficient. While \textsc{rtmba} converges to the optimal policy within tens of episodes, \textsc{dyna} takes approximately 650 episodes to converge, and \textsc{q-learning} takes approximately 22,000. These methods learn in less wall clock time simply because they are able to take many more actions than \textsc{rtmba} in a given amount of time. On an actual robot, it will not be possible to take actions faster than the robot's control frequency, and the poor sample efficiency of these methods will result in longer wall clock learning times as well. In comparison, \textsc{rtmba} learns in fewer actions, meeting our first requirement of sample efficiency even while running at reasonable robot control rates between 10 and 100 Hz.

In addition to enabling real-time learning, another benefit of \textsc{rtmba} is its ability to take advantage of multi-core processors. We ran experiments comparing the performance of \textsc{rtmba} when running on one versus multiple cores.
These experiments were performed on a machine with four 2.6 GHz AMD Opteron processors. Figure~\ref{fig:multivsingle} shows the average reward per episode for these experiments, running at 25 Hz. For comparison, we ran the sequential method using \textsc{mcts} as a planner on the multi-core machine. It had unlimited time for model updates and then planned for 0.04 seconds (the same time given to \textsc{rtmba} for both computations). Since the sequential architecture only has a single thread, it only used a single core even on the multi-core machine. Meanwhile, \textsc{rtmba} utilized three processors with each thread running on its own core. Using the extra processors allowed the parallel version to perform more model updates and planning rollouts between actions than the single core version.
Due to these advantages, the multi-core version performs better than
the single core version, receiving significantly more rewards on every episode ($p < 0.005$). 
In addition, it even performs better than the sequential method on episodes 3 to 14 ($p < 0.01$), even though the 
sequential method is given unlimited time for model updates.

These results demonstrate that the algorithms using our real-time
architecture are able to accomplish both requirements set forth in the
introduction (sample efficiency and real-time action selection), while existing model-free and model-based methods are each only able to accomplish one of the two requirements. 
We have demonstrated that while using approximate planning reduces the time required by model-based methods, they do not reach real-time performance without our parallel architecture.
While agents using \textsc{rtmba} do not learn as much as the sequential methods per action due to the limited time between actions, they still took no more than 5 extra episodes to learn the task. In addition, they were able to learn the task much faster in wall clock time than the sequential algorithms and perform better when run on multiple cores. Next, we look at how the algorithms compare on a task that \emph{requires} real-time actions, where the world will not wait while the agent decides what to do.

\subsection{Autonomous Vehicle}

\begin{figure}[tb]
\begin{center}
\includegraphics[width=0.9\columnwidth]{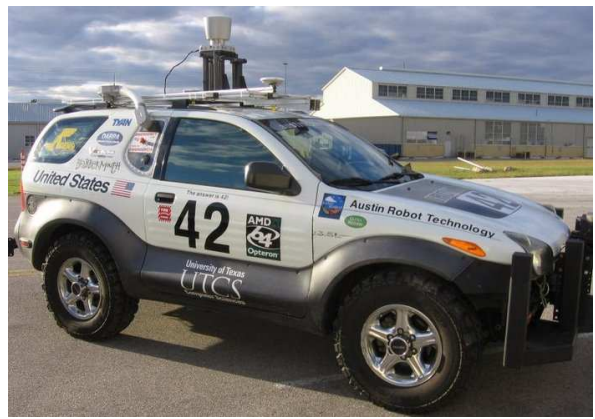}
\end{center}
\vspace{-0.2cm}
\caption{The autonomous vehicle operated by Austin Robot Technology and The University of Texas at Austin.}
\vspace{-0.25cm}
\label{fig:robot}
\end{figure}

Our next task was to control an autonomous vehicle. Here, actions must
be taken in real-time, as the car cannot wait for an action while a
car is stopping in front of it or it approaches a turn in the road.
This task was the main motivator for the creation of \textsc{rtmba}. To
the best of our knowledge, no prior RL algorithm is able to learn in
this domain \emph{in real time}: with no prior data-gathering phase
for training a model.
These experiments take place on the Austin Robot Technology
autonomous vehicle~\cite{JOPHA08-beeson}, 
and on its simulation in ROS stage~\cite{ROS09}.
The vehicle is an Isuzu VehiCross (Figure~\ref{fig:robot}) that has been upgraded to run autonomously
by adding shift-by-wire, steering, and braking
actuators to the vehicle.

Our experiments were to learn to drive the vehicle at
a desired velocity by controlling the pedals. For learning this task,
the RL agent's state was the desired velocity of the vehicle, the
current velocity, and the current position of the brake and
accelerator pedals. Desired velocity was discretized into 0.5 m/s
increments, current velocity into 0.1 m/s increments, and the pedal
positions into tenths of maximum position. The agent's reward at each
step was $-10$ times the error in velocity in m/s. Each episode was
run at 20 Hz (the frequency that the vehicle receives new sensations)
for 10 seconds. The agent had 5 actions: one did nothing (no-op), two
increased or decreased the brake position by 0.1 while setting the
accelerator to 0, and two increased or decreased the accelerator
position by 0.1 while setting the brake position to 0.

The autonomous vehicle software uses ROS~\cite{ROS09} as the underlying middleware. 
We created an RL Interface node that wraps sensor values into \emph{states}, translates \emph{actions} into actuator commands, and generates \emph{reward}.
This node uses a standard set of messages
to communicate with the learning algorithm\footnote{These messages are defined at: \url{http://www.ros.org/wiki/rl_msgs}}, similar to the messages used by \textsc{rl-glue}~\cite{rl-glue}. 
At each time step, it computes the current state and reward and publishes them as a message to the RL agent. The RL agent can then process this information and publish an action message, which the interface will convert into actuator commands.
Whereas the RL agents using \textsc{rtmba} respond with an
action message immediately after receiving the state and reward
message, the sequential methods may have a long delay to complete
model updates and planning before sending back an action message. In
this case, the vehicle continues with all the actuators in their
current positions until it receives a new action message.

\begin{figure}[tbp]
\begin{center}
\includegraphics[width=0.9\linewidth]{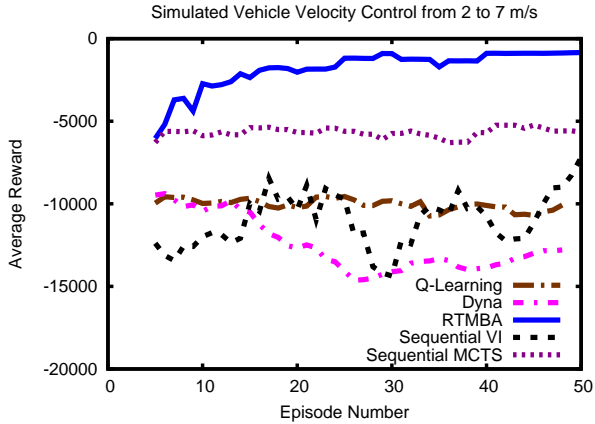}
\end{center}
\vspace{-0.2cm}
\caption{Average rewards of the algorithms controlling the autonomous vehicle in simulation from 2 to 7 m/s. Results are averaged over a 4 episode sliding window.}
\vspace{-0.3cm}
\label{fig:seqcar}
\end{figure}

\begin{figure}[tbp]
\begin{center}
\includegraphics[width=0.9\linewidth]{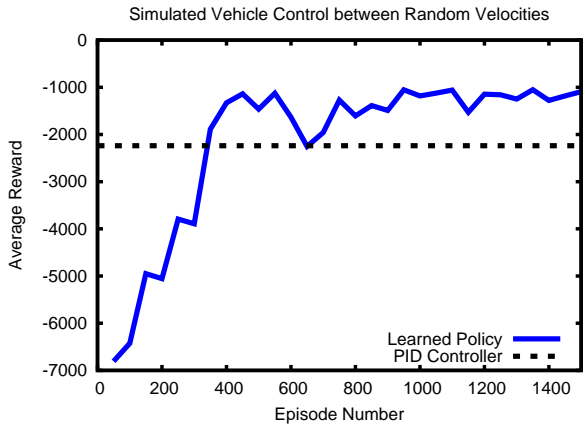}
\end{center}
\vspace{-0.2cm}
\caption{Average rewards of the algorithms controlling the autonomous vehicle in simulation from between random velocities. Results are averaged over a 50 episode sliding window.}
\vspace{-0.3cm}
\label{fig:car-rewards}
\end{figure}

\begin{figure}[tbp]
\begin{center}
\includegraphics[width=0.9\linewidth]{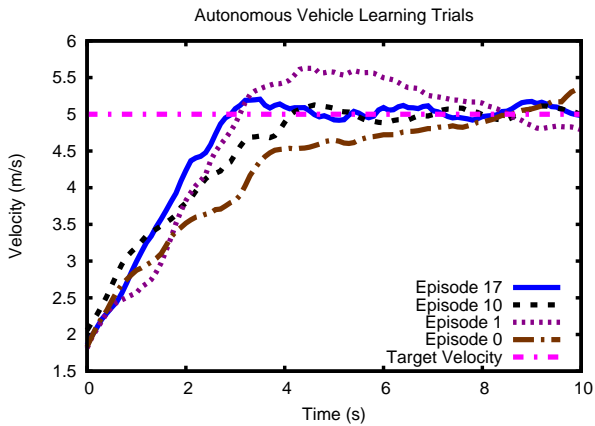}
\end{center}
\vspace{-0.2cm}
\caption{Control profiles of learning trials performed on the physical vehicle.}
\vspace{-0.3cm}
\label{fig:car-trials}
\end{figure}

We ran the first experiment in the ROS stage simulation with the vehicle starting at 2 m/s with a target velocity of 7 m/s. Figure~\ref{fig:seqcar} shows the average rewards per episode for this task. 
Again the model-free methods are not able to learn the task within the given number of episodes.
As before, planning approximately with \textsc{mcts} is better than performing exact planning, but using \textsc{rtmba} is better than either. 
The varying time taken between actions by the sequential methods results in a more difficult learning problem, as the vehicle will have accelerated/decelerated by varying amounts between each action. In only a few minutes, \textsc{rtmba} learns to quickly accelerate to and maintain a velocity of 7 m/s.

Next, we evaluated \textsc{rtmba} on the full velocity control problem, with starting and target velocities selected randomly from between 0 and 11 m/s.
Figure~\ref{fig:car-rewards} shows the reward accrued by the RL agent on each episode in the simulator while learning this task. For comparison, we show the reward that would be received by the PID controller that was previously used for controlling the car's velocity. The previous controller was hand-tuned for performance on the actual car.
The learned controller received more reward than the PID controller after episode 350, which equates to about 1 hour of driving. It was significantly better than the PID controller ($p < 0.005$) after episode 750.

After testing in simulation, we ran learning experiments in real-time on the physical vehicle, learning to drive at 5 m/s from a start of 2 m/s. The velocity curves for a few of the 20 episodes are shown in Figure~\ref{fig:car-trials}. Similar to the simulation results for 2 to 7 m/s, the algorithm learned quickly and was able to accurately track the velocity after 18 episodes (3 minutes of driving).

\section{Related Work}

Batch methods such as experience replay~\cite{lin-expreplay}, fitted Q-iteration~\cite{ErnstGW03}, and \textsc{lspi}~\cite{lagoudakis03leastsquares} improve the sample efficiency of model-free methods by saving experiences and re-using them in periodic batch updates. However, these methods typically run one policy for a number of episodes, stop to perform their batch update, and then repeat. 
Our architecture will also update the model with batches of experience
at a time when the agent is acting faster than it can update the
model. However, \textsc{rtmba} continues taking actions in real-time even while these updates are occurring.

\textsc{dyna}~\cite{Sutton90integratedarchitectures} takes a similar approach to these methods, performing small batch updates between each action. 
The \textsc{dyna-2} framework~\cite{ICML08-silver} extends
\textsc{dyna} to use \textsc{uct} as its planning algorithm, 
combined with permanent and transient memories using linear
function approximation.
This improves the planning
performance of the algorithm, but
the sample efficiency of these
methods still does not meet the requirements for on-line learning laid out in the
introduction.

Deisenroth and Rasmussen~\cite{Deisenroth2009} develop a sample efficient model-based algorithm that uses Gaussian Process regression to compute the model and the policy. It runs in batches, collecting experiences with the current policy before stopping to update its model and plan. 
The algorithm learns to control a physical cart-pole device with few samples, but pauses for 10 minutes of computation after every 2.5 seconds of action. \textsc{rtmba} similarly does batch-type updates to its model, but its parallel architecture allows it to act continually in the domain while performing these updates.

In summary, while there is related work on making model-free methods
more sample-efficient and making model-based methods more reactive,
they all have drawbacks. They either have long pauses in learning to
perform batch updates, or require complete model update or planning
steps between actions. None of these methods accomplish both goals of
being sample efficient and acting continually in real-time, while
\textsc{rtmba} accomplishes both.

\section{Conclusion}

For RL to be practical for continual, on-line learning on a broad range of robotic tasks,
it must both (1) be sample-efficient and (2) learn while taking actions continually
in real-time. 
This paper introduces a novel parallel architecture for model-based RL
that is the first to enable an agent to act in real-time while
maintaining the sample efficiency of model-based RL. It uses sample-based
approximate planning and performs model
learning and planning in parallel threads, while a third thread
returns actions at a rate dictated by the task.
In addition, \textsc{rtmba} enables RL algorithms to take advantage of the
multi-core processors available on many robotic platforms. 
Our experiments, in simulation and on a real robot, demonstrate that
this architecture is necessary for learning on robots that require 
fast real-time actions.
Our ongoing
research agenda includes testing \textsc{rtmba} on other robotic
platforms, as well as testing other model learning and \textsc{mcts}
planning algorithms within the framework.

\section*{Code}

Source code for the real-time architecture, algorithms, and experiments described in this paper are available
as part of a ROS repository available at: 
\begin{small}\url{http://www.ros.org/wiki/rl-texplore-ros-pkg}\end{small}
The architecture and the \textsc{texplore} algorithm used for model learning are available 
in that repository in the \textsc{rl\_agent} package available at: 
\begin{small}\url{http://www.ros.org/wiki/rl_agent}\end{small}
In addition, the mountain car task and a simplified version of the autonomous vehicle velocity control task are 
also available in the repository in the \textsc{rl\_env} package available at: 
\begin{small}\url{http://www.ros.org/wiki/rl_env}\end{small}
Finally, the definitions of the ROS messages used to communicate between agent and environment are available
in the repository in the \textsc{rl\_msgs} package available at: 
\begin{small}\url{http://www.ros.org/wiki/rl_msgs}\end{small}

The experimental results presented in this paper can be reproduced easily. For example, to run \textsc{texplore} controlling the simulated vehicle from 2 to 7 m/s, with the \textsc{rl\_experiment} package installed, type:
\begin{small}\texttt{rosrun rl\_experiment experiment --agent texplore --env car2to7 --actrate 20}\end{small}

\section*{Acknowledgments}

\thanks{This work has taken place in the Learning Agents Research
Group (LARG) at the Artificial Intelligence Laboratory, The University
of Texas at Austin. LARG research is supported in part by grants from
the National Science Foundation (IIS-0917122), ONR (N00014-09-1-0658),
and the Federal Highway Administration (DTFH61-07-H-00030).}

\bibliographystyle{IEEEtran}
\bibliography{../bib/rl}
\end{document}